\documentclass[a4paper, 10 pt, conference]{ieeeconf}  %

\IEEEoverridecommandlockouts                              %
\makeatletter
\let\NAT@parse\undefined
\makeatother

\usepackage[dvipsnames]{xcolor}

\newcommand*\linkcolours{ForestGreen}

\usepackage[left=2cm,right=2cm,top=2cm,bottom=2cm]{geometry}
\usepackage{times}
\usepackage{todonotes}
\usepackage{graphicx}
\usepackage{amssymb}
\usepackage{algorithm}
\usepackage{algpseudocode}
\usepackage{gensymb}
\usepackage{amsmath}
\usepackage{subcaption}
\usepackage{breakurl}
\usepackage{adjustbox}
\usepackage{array}
\usepackage[normalem]{ulem}
\useunder{\uline}{\ul}{}
\usepackage{booktabs}

\usepackage{url,hyperref}
\hypersetup{
colorlinks,
linkcolor=\linkcolours,
citecolor=\linkcolours,
filecolor=\linkcolours,
urlcolor=\linkcolours}

\usepackage{xcolor}

\usepackage[labelfont={bf},font=small]{caption}
\usepackage[none]{hyphenat}

\usepackage{mathtools, cuted}

\usepackage[noadjust, nobreak]{cite}

\usepackage{parskip}
\usepackage{tabularx}
\usepackage{amsmath}
\usepackage{multirow}

\usepackage{float}

\usepackage{pifont}%

\newcolumntype{Y}{>{\centering\arraybackslash}X}

\usepackage[]{placeins}

\usepackage{placeins}

\usepackage{tikz}

\usepackage[framemethod=tikz]{mdframed}

\usepackage{afterpage}

\usepackage{stfloats}

\usepackage{atbegshi}
\newcommand{\handlethispage}{}
\newcommand{\discardpagesfromhere}{\let\handlethispage\AtBeginShipoutDiscard}
\newcommand{\keeppagesfromhere}{\let\handlethispage\relax}
\AtBeginShipout{\handlethispage}

\usepackage{comment}

\title{\LARGE \bf
Skill Issues: An Analysis of CS:GO Skill Rating Systems
}

\author{Mikel Bober-Irizar, Naunidh Dua \& Max McGuinness*\thanks{*All authors contributed equally to the work.}\\
University of Cambridge
}

\newcolumntype{R}[2]{%
    >{\adjustbox{angle=#1,lap=\width-(#2)}\bgroup}%
    l%
    <{\egroup}%
}

\newcommand{\sss}[1]{
\vspace{2mm}\subsubsection{\textbf{#1}}
}

\begin{document}

\thispagestyle{plain}
\pagestyle{plain}
\maketitle

\newcommand{\sensitivityanalysisfigure}[1]{
\begin{figure*}[#1]
    \begin{subfigure}{\linewidth}
    \hspace{-0.025\linewidth}
        \includegraphics[width=1.05\textwidth]{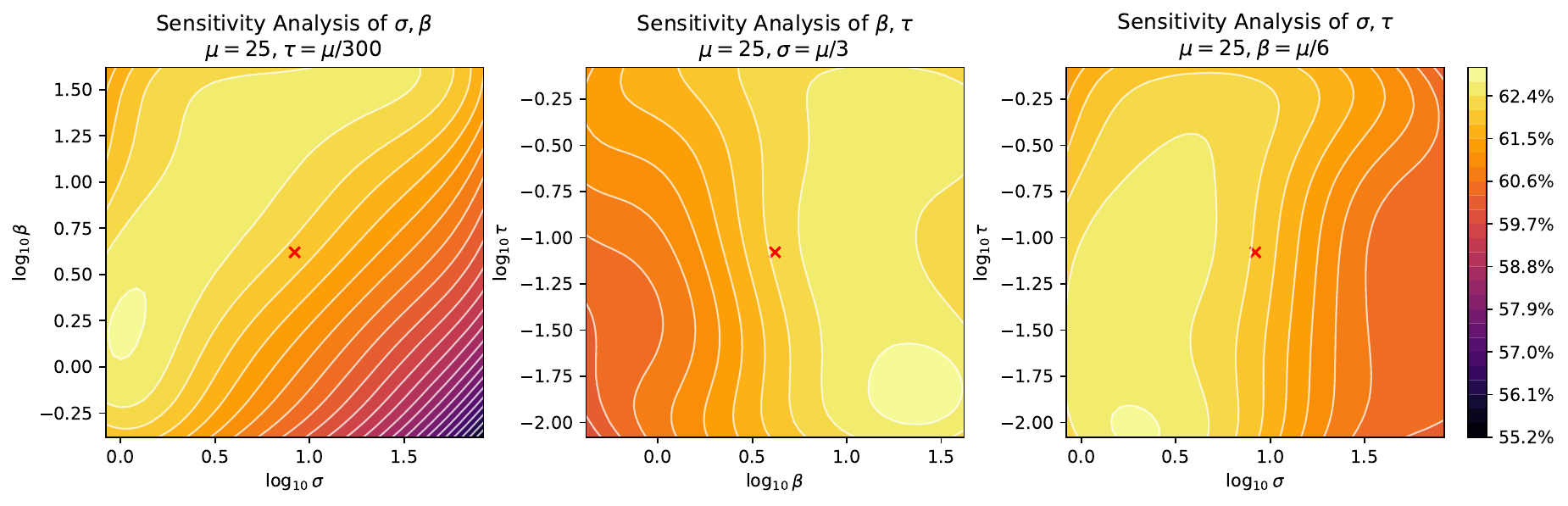}
        \vspace{-6mm}
        \caption{\textbf{TrueSkillEmulator}}
    \end{subfigure}
    \begin{subfigure}{\linewidth}
    \vspace{2mm}
    \hspace{-0.025\linewidth}
          \includegraphics[width=1.05\textwidth]{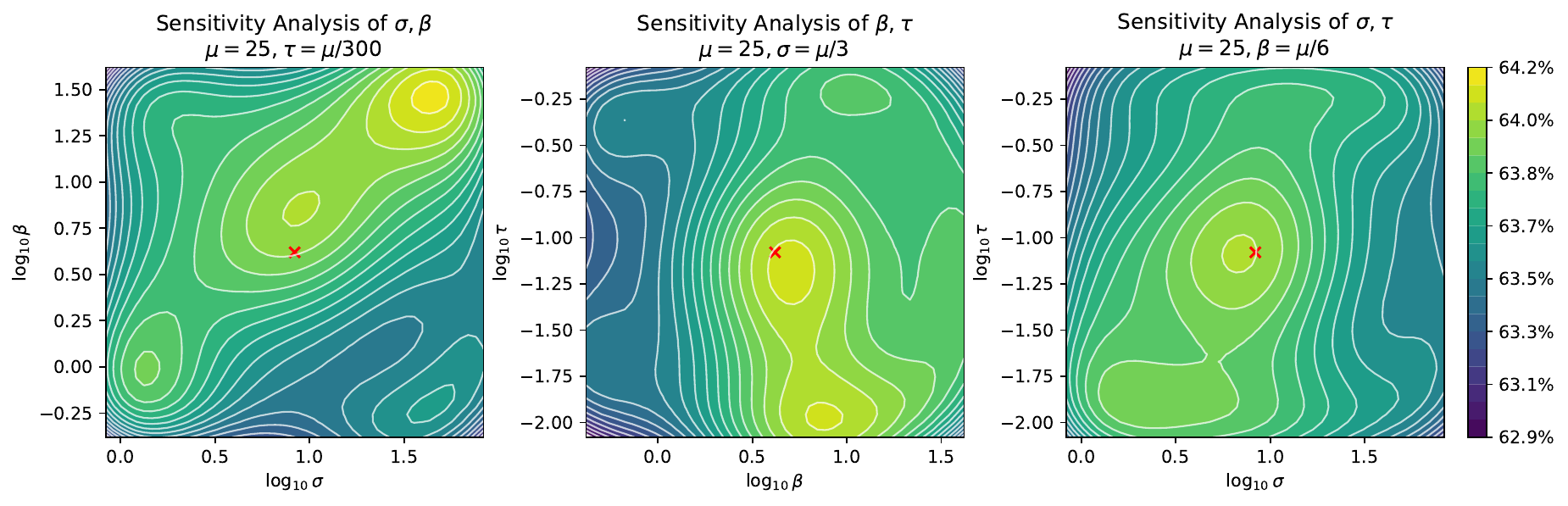}
          \vspace{-6mm}
          \caption{\textbf{TrueSkillPlayersEmulator}}
    \end{subfigure}
  \caption{Sensitivity analysis for our TrueSkill emulators, after 2000 training matches selected by the \textbf{LikeliestDraw (\ref{eq:AFdraw})} acquisition function. The red \textbf{\textcolor{red}{$\times$}} shows the default TrueSkill parameters, with the plot being $\pm 1$ order of magnitude. Note that the scale in (a) is a much larger range than in (b).}
  \label{fig:ts-sensitivity}
\end{figure*}
}

\newcommand*\rot{\multicolumn{1}{R{45}{1em}}}%
\newcommand{\maintable}[1]{

\setlength{\tabcolsep}{4pt}
\begin{table*}[#1]
\centering
\begin{tabular}{crcccccccc}
\toprule

\textbf{Training} & \textbf{Emulator} & \rot{Random}& \rot{MostSeen (\ref{sec:AFother})}& \rot{LeastSeen (\ref{AFleastseen})}& \rot{LikeliestWin (\ref{sec:AFother})}& \rot{LikeliestDraw (\ref{eq:AFdraw})}& \rot{CrossEntropy (\ref{eq:AFCE})}& \rot{Weighted (\ref{eq:AFweighted})}&  \rot{TSQuality (\ref{sec:AFother})}\\
\midrule
\multirow{ 8}{*}{500 matches}
& \textbf{Random} & \colorbox{green!0}{50.1\%}&\colorbox{green!0}{50.0\%}&\colorbox{green!0}{50.0\%}&\colorbox{green!0}{49.9\%}&\colorbox{green!0}{\textbf{50.1\%}}&\colorbox{green!0}{50.0\%}&\colorbox{green!0}{50.0\%}&\colorbox{green!0}{-} \\
& \textbf{WinRate} & \colorbox{green!0}{58.8\%}&\colorbox{green!0}{54.8\%}&\colorbox{green!0}{58.9\%}&\colorbox{green!0}{58.1\%}&\colorbox{green!0}{58.3\%}&\colorbox{green!0}{\textbf{59.0\%}}&\colorbox{green!0}{59.0\%}&\colorbox{green!0}{-} \\
& \textbf{Elo} & \colorbox{green!0}{59.3\%}&\colorbox{green!0}{55.4\%}&\colorbox{green!0}{59.7\%}&\colorbox{green!0}{56.1\%}&\colorbox{green!0}{\textbf{59.8\%}}&\colorbox{green!0}{59.4\%}&\colorbox{green!0}{59.5\%}&\colorbox{green!0}{-} \\
& \textbf{Glicko2} & \colorbox{green!1}{60.1\%}&\colorbox{green!0}{58.3\%}&\colorbox{green!0}{59.4\%}&\colorbox{green!0}{58.8\%}&\colorbox{green!6}{60.5\%}&\colorbox{green!3}{60.3\%}&\colorbox{green!15}{\textbf{61.2\%}}&\colorbox{green!0}{-} \\
& \textbf{TrueSkill} & \colorbox{green!0}{59.1\%}&\colorbox{green!0}{57.4\%}&\colorbox{green!0}{59.0\%}&\colorbox{green!0}{58.2\%}&\colorbox{green!0}{59.2\%}&\colorbox{green!0}{58.9\%}&\colorbox{green!4}{\textbf{60.4\%}}&\colorbox{green!0}{56.5\%} \\
& \textbf{TSPlayers} & \colorbox{green!0}{59.6\%}&\colorbox{green!0}{56.3\%}&\colorbox{green!6}{60.6\%}&\colorbox{green!0}{56.5\%}&\colorbox{green!11}{60.9\%}&\colorbox{green!12}{61.0\%}&\colorbox{green!26}{\textbf{62.1\%}}&\colorbox{green!0}{57.8\%} \\
\cmidrule(lr){2-10}
& \textbf{Average} & \colorbox{green!0}{59.4\%}&\colorbox{green!0}{56.4\%}&\colorbox{green!0}{59.5\%}&\colorbox{green!0}{57.5\%}&\colorbox{green!0}{59.7\%}&\colorbox{green!0}{59.7\%}&\colorbox{green!5}{\textbf{60.4\%}}&- \\
\midrule
\multirow{ 8}{*}{1000 matches}
& \textbf{Random} & \colorbox{green!0}{50.1\%}&\colorbox{green!0}{50.1\%}&\colorbox{green!0}{\textbf{50.1\%}}&\colorbox{green!0}{49.9\%}&\colorbox{green!0}{49.9\%}&\colorbox{green!0}{50.0\%}&\colorbox{green!0}{50.0\%}&\colorbox{green!0}{-} \\
& \textbf{WinRate} & \colorbox{green!6}{60.5\%}&\colorbox{green!0}{55.9\%}&\colorbox{green!3}{60.3\%}&\colorbox{green!0}{59.9\%}&\colorbox{green!0}{59.4\%}&\colorbox{green!10}{\textbf{60.8\%}}&\colorbox{green!8}{60.7\%}&\colorbox{green!0}{-} \\
& \textbf{Elo} & \colorbox{green!12}{61.0\%}&\colorbox{green!0}{56.6\%}&\colorbox{green!11}{60.9\%}&\colorbox{green!0}{57.8\%}&\colorbox{green!17}{61.4\%}&\colorbox{green!12}{61.0\%}&\colorbox{green!27}{\textbf{62.2\%}}&\colorbox{green!0}{-} \\
& \textbf{Glicko2} & \colorbox{green!17}{61.4\%}&\colorbox{green!0}{60.1\%}&\colorbox{green!3}{60.3\%}&\colorbox{green!0}{59.5\%}&\colorbox{green!24}{62.0\%}&\colorbox{green!14}{61.2\%}&\colorbox{green!29}{\textbf{62.4\%}}&\colorbox{green!0}{-} \\
& \textbf{TrueSkill} & \colorbox{green!9}{60.8\%}&\colorbox{green!0}{59.1\%}&\colorbox{green!0}{59.7\%}&\colorbox{green!0}{59.5\%}&\colorbox{green!10}{60.9\%}&\colorbox{green!1}{60.1\%}&\colorbox{green!22}{\textbf{61.8\%}}&\colorbox{green!0}{58.2\%} \\
& \textbf{TSPlayers} & \colorbox{green!9}{60.7\%}&\colorbox{green!0}{57.7\%}&\colorbox{green!10}{60.8\%}&\colorbox{green!0}{57.5\%}&\colorbox{green!34}{62.8\%}&\colorbox{green!22}{61.8\%}&\colorbox{green!39}{\textbf{63.2\%}}&\colorbox{green!0}{60.1\%} \\
\cmidrule(lr){2-10}
& \textbf{Average} & \colorbox{green!10}{60.9\%}&\colorbox{green!0}{57.9\%}&\colorbox{green!5}{60.4\%}&\colorbox{green!0}{58.9\%}&\colorbox{green!16}{61.3\%}&\colorbox{green!12}{61.0\%}&\colorbox{green!25}{\textbf{62.0\%}}&- \\
\midrule
\multirow{ 8}{*}{2000 matches}
& \textbf{Random} & \colorbox{green!0}{\textbf{50.1\%}}&\colorbox{green!0}{50.1\%}&\colorbox{green!0}{50.0\%}&\colorbox{green!0}{50.0\%}&\colorbox{green!0}{50.0\%}&\colorbox{green!0}{50.1\%}&\colorbox{green!0}{50.0\%}&\colorbox{green!0}{-} \\
& \textbf{WinRate} & \colorbox{green!23}{61.9\%}&\colorbox{green!0}{57.0\%}&\colorbox{green!25}{62.0\%}&\colorbox{green!15}{61.3\%}&\colorbox{green!1}{60.2\%}&\colorbox{green!28}{\textbf{62.3\%}}&\colorbox{green!15}{61.3\%}&\colorbox{green!0}{-} \\
& \textbf{Elo} & \colorbox{green!28}{62.3\%}&\colorbox{green!0}{58.0\%}&\colorbox{green!24}{62.0\%}&\colorbox{green!0}{59.8\%}&\colorbox{green!28}{62.3\%}&\colorbox{green!26}{62.2\%}&\colorbox{green!34}{\textbf{62.8\%}}&\colorbox{green!0}{-} \\
& \textbf{Glicko2} & \colorbox{green!32}{62.6\%}&\colorbox{green!17}{61.4\%}&\colorbox{green!27}{62.2\%}&\colorbox{green!6}{60.5\%}&\colorbox{green!37}{63.0\%}&\colorbox{green!38}{63.1\%}&\colorbox{green!39}{\textbf{63.1\%}}&\colorbox{green!0}{-} \\
& \textbf{TrueSkill} & \colorbox{green!28}{62.2\%}&\colorbox{green!10}{60.8\%}&\colorbox{green!23}{61.9\%}&\colorbox{green!11}{60.9\%}&\colorbox{green!27}{62.2\%}&\colorbox{green!22}{61.8\%}&\colorbox{green!35}{\textbf{62.9\%}}&\colorbox{green!4}{60.4\%} \\
& \textbf{TSPlayers} & \colorbox{green!22}{61.8\%}&\colorbox{green!0}{59.2\%}&\colorbox{green!8}{60.7\%}&\colorbox{green!0}{59.4\%}&\colorbox{green!49}{63.9\%}&\colorbox{green!32}{62.6\%}&\colorbox{green!50}{\textbf{64.1\%}}&\colorbox{green!36}{62.9\%} \\
\cmidrule(lr){2-10}
& \textbf{Average} & \colorbox{green!27}{62.2\%}&\colorbox{green!0}{59.3\%}&\colorbox{green!21}{61.8\%}&\colorbox{green!4}{60.4\%}&\colorbox{green!28}{62.3\%}&\colorbox{green!29}{62.4\%}&\colorbox{green!35}{\textbf{62.8\%}}&- \\
\bottomrule

\end{tabular}
\caption{Evaluation accuracy after 500 to 2000 training matches (roughly 1-4 matches per team), for each Emulator and Aquisition Function combination. Each point is an average of 100 runs, with a epistemic confidence interval $<0.1\%$. The best acquisition function for each emulator is bolded. Average values exclude the Random Emulator.}.
\label{table:main}
\end{table*}
}

\newcommand{\accuracytraininggraph}[1]{
\begin{figure*}[#1]
\includegraphics[width=\textwidth]{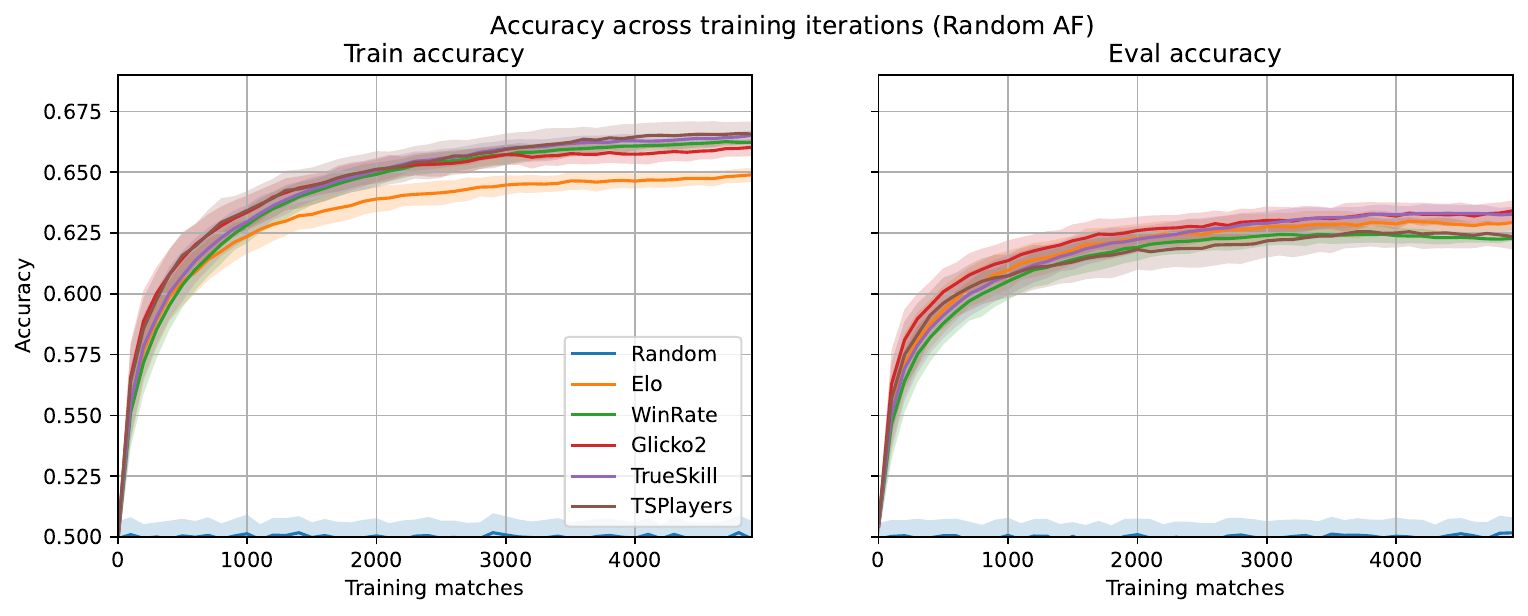}
\includegraphics[width=\textwidth]{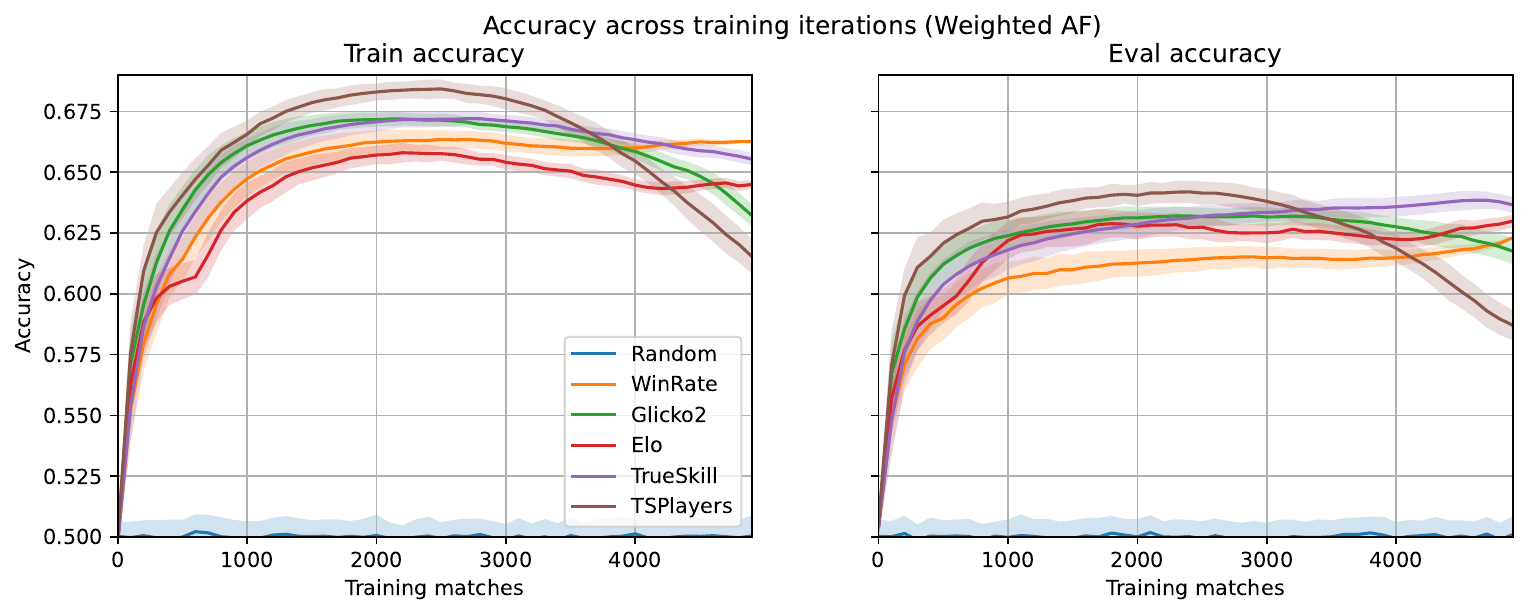}
\caption{The training and evaluation accuracy across emulators, using a Random and Weighted \ref{eq:AFweighted} AF. As the dataset is exhausted, both acquisition functions train on all matches, in a different order. Error bars show $\pm 1 \sigma$ of \textbf{aleatoric} uncertainty; the variance between individual runs of the Simulator.}
\label{fig:curve}
\end{figure*}
}

\begin{abstract}

The meteoric rise of online games has created a need for accurate \textit{skill rating} systems for tracking improvement and fair matchmaking. Although many skill rating systems are deployed, with various theoretical foundations, less work has been done at analysing the real-world performance of these algorithms. In this paper, we perform an empirical analysis of Elo, Glicko2 and TrueSkill through the lens of surrogate modelling, where skill ratings influence future matchmaking with a configurable acquisition function. We look both at overall performance and data efficiency, and perform a sensitivity analysis based on a large dataset of Counter-Strike: Global Offensive matches.

\end{abstract}

\section{Introduction}

Counter-Strike: Global Offensive (CS:GO) is a multiplayer first-person shooter game where players work in teams of 5v5 to fight over objectives, and ultimately try to win a match of up to 30 rounds. As with many modern games, gameplay is focused around a central ``competitive matchmaking'' mode, where two teams of five with similar skill are pitted against each other, and the outcome of each match is used to update the players' skill ratings.

With millions of competitive matches played each day \cite{csgo-player-count}, having accurate skill ratings for each player and team is fundamental to producing fair matchups and allowing players to progress as their skill improves. The ever-rising popularity of matchmaking in multiplayer video games has shown that this format makes for an engaging experience. 

In CS:GO, the developer \textit{Valve} uses an unpublished variant of the Glicko2 rating system \cite{glicko2, csgo-glicko2}; the perceived inaccuracies of the skill rating system are a major point of contention in the community. The term ``elo hell'' is commonly used by players who feel that their skill rating doesn't reflect their real ability \cite{elohell}, and many players opt to use alternative matchmaking services such as FACEIT (which uses the Elo system \cite{elo, faceit-elo}) to form ratings instead. While the benefits of accurate skill ratings are clear, little comparative work has been done to understand the performance of each system in CS:GO.

In this work, we take a look at several systems available for skill ratings (including Elo and Glicko2), and apply it to a large dataset of professional CS:GO matches. As well as comparing the skill rating systems themselves, we explore the effect of different matchmaking algorithms on the accuracy of skill ratings. We achieve this through a surrogate modelling environment where the matchmaking system chooses which match to be played next (and therefore to be used for updating skill ratings before the next match), while imputing real data for the outcomes of matches. This allows us to quantify the effect of the circular dependency between skill ratings and matchmaking algorithms, as opposed to measuring the behaviour of each skill rating system on a static set of past matches.

\section{Related Work}

The history of skill rating systems for games is a rich one. The Elo system, first introduced in the 1950s, is widely used in many sports and remains the defining system the chess world accepts. The system assumes that each player has a fixed skill rating, and the probability of each player winning the match is a function of the difference in rating between the two players \cite{elo}. It was the first rating system developed that modelled the players' skill level probabilistically.

In 1995, Mark Glickman created the Glicko system, specifically to improve upon issues he saw with the Elo system \cite{glicko, glicko2}. Glicko adds a confidence parameter \textit{RD}, that is a measure of the system's confidence in its estimate of skill. This allows changes in skill rating to also be dependent on this confidence parameter, thus introducing an idea of ``information gained" by a particular match being played. Glicko2 adds an additional parameter $\sigma$, that measures the player's fluctuation in skill.

Herbrich, \textit{et al.\ } from Microsoft Research introduced TrueSkill in 2007, based on Bayesian inference \cite{trueskill}. TrueSkill address two concerns of online team based matchmaking:
\begin{enumerate}
    \item Match outcomes are team-based, but a skill rating for individuals is desired.
    \item Some games have multiple `teams' playing, and the match outcomes are not binary win/loss, (e.g. ``free-for-all (FFA) deathmatch" matches)
\end{enumerate}
TrueSkill uses Gaussian Processes (GPs) via a Factor Graph and Message Passing approach to model the players' skill and perform matchmaking. Microsoft analysed the performance of TrueSkill on a dataset collected in the beta testing of Halo 2, and found that TrueSkill substantially outperforms Elo \cite{trueskill}. Microsoft published a significant upgrade, TrueSkill 2, in 2018 \cite{trueskill2}. Trueskill 2 is designed to address deficiencies in TrueSkill by analysing game-specific metrics such as player experience, the player's individual performance in the match, tendency to quit, and skill in other game modes  to assist with matchmaking.

Some evaluation of prediction performance has been previously studied. Dehpanah \textit{et al.\ }analysed Elo, Glicko and TrueSkill on a set of 100,000  matches from PlayerUnknown’s Battlegrounds \cite{related_rating_eval}. They found that incorporation of new players into the model proved tricky, and deteriorated the performance of these rating systems. 

Makarov \textit{et al.\ }analysed the performance of rating systems on two \textit{Valve} games, focusing on Dota 2 as well as CS:GO. They analysed TrueSkill on CS:GO games and observed a 62\% accuracy \cite{csgo_dota_prediction}. Both approaches analysed a static dataset of past matches, in contrast with our approach.

\section{Methods}

\subsection{Framework}

For the experiments in this paper, we implemented a extensible Python library called \textbf{skillbench}, which allowed us to implement each component for comparison following a consistent interface. Skill rating systems such as TrueSkill are implemented as \textbf{Emulators}, with matchmaking algorithms implemented as \textbf{Acquisition Functions}. The overall training and evaluation is governed by a \textbf{Simulator}, which simulates the playing of games (as chosen by the acquisition function) based on the match dataset, and trains and evaluates the emulator.

\begin{figure}[H]
\centering
\includegraphics[width=\hsize]{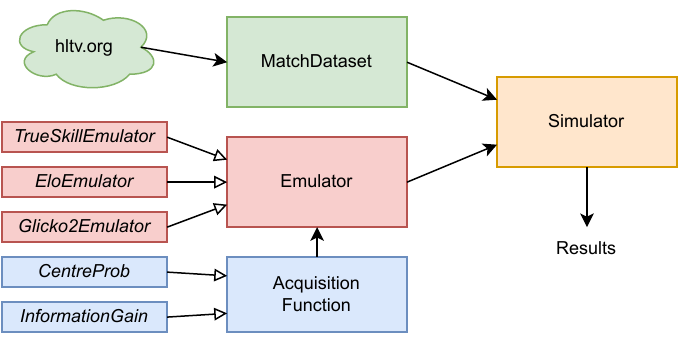}
\caption{The architecture of the \textbf{skillbench} library. \textit{Emulator}s and their \textit{Acquisition Function}s are implemented as modular components following a common interface. A \textit{Simulator} takes an \textit{Emulator}, trains it on a training \textit{MatchDataset}, and evaluates it on an evaluation \textit{MatchDataset}.}
\label{fig:arch}
\end{figure}

Figure \ref{fig:arch} shows the overall architecture of the library. In the following sections, we talk about each component in much more detail.

\subsection{Simulator}

In lieu of a game environment populated by real players, we simulate the process of chosen teams competing against one another by selecting records from a dataset of historical matches (Section \ref{sec:data}). The key constraint to this approach is that the model's choice of matchups is limited to those present in the dataset - these limitations are discussed further in Section \ref{sec:limitations}.

In our implementation, we characterise the simulator as being responsible for the generic training process of emulators. That is to say, it is within the \textit{Simulator} class that a matchup is chosen according to its acquisition score (computed by an \textit{Acquisition Function}); the result of the match is then simulated (by popping a random result from the set of real matches between that pair of teams); before finally the \textit{Emulator} is informed of the result and it is fit into its internal model.

At each iteration, the Simulator chooses 25 matches from the remaining training pool at random, and the emulator is fed whichever match has the highest score according to an \textit{Acquisition Function}. After a given number of training matches, we evaluate the Emulator's predictions according to its accuracy against historical results. For external validity, we can split our dataset and make use of separate ``training" and ``testing" Simulators.

We summarise this \textit{Simulator}-led training process below.

\begin{algorithm}
\caption{Simulator.Train (Emulator \textbf{E}, Acquisition Function \textbf{AF})}\label{alg:simulator_train}
\begin{algorithmic}
\Require{loaded match dataset \textbf{M}.}
\For{$i$ in num\_evals}
\State $\lambda$ $\gets$ 25 random matches from M;
\State m* $\gets$ $\text{argmax}_{m \in \lambda}$[AF(E, m.T1, m.T2)];
\State M.pop(m*);
\State E.fit(m*);
\EndFor
\end{algorithmic}
\end{algorithm}

\begin{algorithm}
\caption{Simulator.Eval (Emulator \textbf{E})}\label{alg:simulator_eval}
\begin{algorithmic}
\Require{loaded match dataset \textbf{M}.}
\State $\mu$ $\gets$ 0;
\State $\nu$ $\gets$ 0;
\For{m in M}
\If{m.result $\neq$ draw}
\State p $\gets$ E.predict(m.T1, m.T2);
\If{p$>$0.5 \& m.winner = m.T1}
\State $\mu$ $\gets$ $\mu+1$;
\EndIf
\State $\nu$ $\gets$ $\nu+1$;
\EndIf
\EndFor
\State \Return $\mu/\nu$
\end{algorithmic}
\end{algorithm}

\subsection{Emulators}

We model each skill rating system as an \textbf{Emulator}, which can ``emulate" (predict) the outcome of a match between two teams as a probability. Each emulator can also be updated based on the outcome of previous matches.

We implement five emulators within our framework, which are briefly described below.

\sss{WinRate}

WinRate is a baseline naïve emulator which we introduce in this work, as a point of comparison. We keep track of each team's proportion of matches won so far, $w(T)$, and then calculate a `probability' of $A$ winning a match against $B$:
$$E[A|B] = \frac{1 + w(A) - w(B)}{2}$$

One of the shortcomings of this approach is that a team's win rate is biased by the skill of the teams they have played against; this is addressed by the other emulators.

\sss{Elo}

Our Elo implementation is defined as follows:

Parameters: $k$, representing the `k-factor' and $\mu$, the starting rating given to new players.

\begin{equation}\label{eq:EloEmu}
\begin{split}
E[A|B] = \frac{1}{1 + 10^{\frac{R_B-R_A}{400}}}
\end{split}
\end{equation}

Equation \ref{eq:EloEmu} represents the expected score (probability of winning) of team $A$ in a match against team $B$, where $R_T$ is the rating of team $T$.

To update the skill ratings, we multiply the difference between the actual outcome and the predicted outcome by $k$, and adjust the rating by this value.

\sss{Glicko2}

The Glicko2 implementation has the following parameters:

\begin{itemize}
    \item $\mu$ = The default rating of player.
    \item $\phi$ = The rating deviation (RD) of a player.
    \item $\sigma$ = The player's skill volatility.
    \item $\tau$ = The system constant, which dictates the volatility over time.
\end{itemize}

The exact specifics of how to calculate the rating algorithm are detailed in \cite{glicko2}; we provide an overview here.

\begin{enumerate}
    \item Compute the estimated variance ($v$) of the team's rating based solely on the game outcome.
    \item Compute the estimated improvement ($\Delta$) in rating by comparing the current rating to the rating based on the game outcome.
    \item Iteratively compute the new volatility.
\end{enumerate}

\vspace{-5mm}
\begin{equation}\label{eq:GlickoV}
\begin{split}
v = \left[ g(\phi')^2 E(\mu, \mu', \phi')\cdot (1-E(\mu,\mu',\phi')) \right]^{-1}
\end{split}
\end{equation}

and

\vspace{-5mm}
\begin{equation}\label{eq:GlickoD}
\begin{split}
\Delta = v g(\phi') ( s - E(\mu,\mu',\phi')) 
\end{split}
\end{equation}

where $\mu$' $\phi$' represent the rating and RD of the opponent, and s is the actual outcome.

\sss{TrueSkill}

The TrueSkill implementation has the following parameters \cite{trueskill, trueskillmath}:

\begin{itemize}
    \item $\mu$ = The rating of player.
    \item $\sigma$ = The player's skill volatility.
    \item $\beta$ = The skill class width, if a player has $\beta$ rating higher than another, then the player has an 80\% chance to win.
    \item $\tau$ = The additive dynamics factor, the square of which is added to the player's variance on each skill update. This ensures that $\sigma$ does not converge to 0, and thus a player's skill rating never becomes static.
\end{itemize}

TrueSkill is a Gaussian Process over teams (or players), that is a joint distribution of infinitely many Gaussians. The TrueSkill algorithm attempts to minimise the Kullback-Leibler (K-L) Divergence between a 3D truncated Gaussian (created by the performance of the two teams) and the approximation of it. 

A deep dive in the maths behind TrueSkill is given in \cite{trueskillmath}.

\sss{TrueSkillPlayers} Similar to TrueSkill, but rather than tracking a per-team skill rating, each player's skill rating is tracked individually. We take advantage of TrueSkill's unique native support for any type of match, to update all 10 skill ratings for each player in one go.

We hypothesise that this could bring benefits, as it allows players to `bring their skill ratings with them' when they occasionally move between teams or form new teams. This is advantageous to the approach taken in other rating systems, where a team needs to retain three players (a `core') in order to keep its rating.

\subsection{Acquisition Functions (AFs)}
One of the goals of this analysis is to determine how to select teams for matches that provide the most information to the skill rating system - i.e. how can we acquire all teams' skill ratings in as few games as possible? To do this, we use the notion of an \textbf{acquisition function} (AF) from surrogate modelling: a function which determines which data point to sample next to update the model. In this case, the data points are match outcomes between some pair of teams, and the model is the skill rating emulator. This can be thought of as an analogue to the matchmaking system which is used in practice for selecting matchups.

An acquisition function provides a heuristic quantification for how valuable any given match may be in the training of an emulator, according to:
\begin{itemize}
\item the emulator's internal state;
\item which teams are involved in the match.
\end{itemize}
More formally, we can define an acquisition function as:
$$\text{AF}: \text{Emulator} \times \text{Team} \times \text{Team} \rightarrow \mathbb{R}.$$
We will now discuss several approaches to designing an acquisition function.

\sss{Expected Improvement}

One particularly prevalent form of acquisition is Expected Improvement (EI), which adopts a greedy strategy to locate the global minimum in a search space. It achieves this by selecting the point that has the greatest probability of being lower than our current best estimate, as predicted by a surrogate model such as a Gaussian Process (GP):
\begin{equation}\label{eq:EI}
\mathbb{E}[u(x)|x,\mathcal{F}] = \mathbb{E}[\max(0, s_\mathcal{F}(x_*)-s_\mathcal{F}(x))|x,\mathcal{F}],
\end{equation}
where $\mathcal{F}$ is our search space ($\mathcal{F}:X \rightarrow Y$), $s_\mathcal{F}$ is our surrogate model approximating $\mathcal{F}$, and $x_*$ is our current best estimate for $\mathcal{F}$'s global minimum.

To translate this approach to the domain of CS:GO skill estimation, one could make the following correspondence:
\begin{table}[H]
\begin{tabularx}{\hsize}[t]{rl}
\toprule
\textbf{} & \textbf{CS:GO Correspondence} \\ \midrule
$\mathcal{F}$ & \begin{tabular}[c]{@{}l@{}}Our emulator's mean error rate at predicting match outcomes,\\ after learning of a particular matchup (i.e. $\mathcal{F}:X \rightarrow Y$).\end{tabular} \\
X & Possible matchups (i.e. Team $\times$ Team). \\
Y & Our emulator's mean error rate at predicting match outcomes. \\
$x_*$ & The most useful matchup according to $s_\mathcal{F}$. \\ \bottomrule
\end{tabularx}
\vspace{-3mm}
\end{table}
However, there is an issue with this setup. The basic formulation of EI (equation \ref{eq:EI}) models the function $\mathcal{F}$ and search domain $X$ as static, when in reality they should depend on our emulator's internal state. In other words, over the course of training, we don't want to find the most useful single matchup, but rather the most useful \textit{collection} of matchups to condition the emulator, each iteration building on our existing collection.

This alters the premise of the Expected Improvement environment - we now have a dynamic search space, which depends on the history of matches $x_{t-1}$ that have already been observed at that timestep:

\begin{table}[H] 
\begin{tabularx}{\hsize}[t]{rl}
\toprule
\textbf{} & \textbf{CS:GO Correspondence} \\ \midrule
$\mathcal{F}_{x_t}$ & \begin{tabular}[c]{@{}l@{}l@{}}Our emulator's mean error rate at predicting match outcomes,\\ after learning of a particular matchup, given match history \\ $x_{t-1}$ (i.e. $\mathcal{F}_{x_t}:X_{x_t} \rightarrow Y$).\end{tabular} \\
$X_{x_t}$ & \begin{tabular}[c]{@{}l@{}} Possible collections of matchups, each extending $x_{t-1}$ by \\ one (i.e. \{$x_{t-1}$\} $\times$ Team $\times$ Team). \end{tabular} \\
Y & Our emulator's mean error rate at predicting match outcomes. \\ 
$x_{t-1}$ & \begin{tabular}[c]{@{}l@{}} The most useful matchup collection according to \\ $s_{\mathcal{F}_{x_{t-1}}}: X_{x_{t-1}} \rightarrow Y$. \end{tabular} \\
\bottomrule
\end{tabularx}
\vspace{-3mm}
\end{table}

Being more explicit with regards to timesteps, we can now reformulate the Expected Improvement equation as:
\begin{equation}\label{eq:EI2}
\begin{split}
\mathbb{E}[u(x)|x,\mathcal{F}] = \mathbb{E}[\max(0, s_{\mathcal{F}_{x_{t-1}}}(x_{t-1}) - s_{\mathcal{F}_{x_t}}(x)) | x, \mathcal{F}].
\end{split}
\end{equation}

\sss{Cheater's AF}
With this EI environment, we can immediately notice a hypothetically optimal (greedy) choice of acquisition function, by making our choice of $s_\mathcal{F}$ as inherently close to $\mathcal{F}_{x_t}$ as possible. In practical terms, this means ``cheating'' and computing $s_\mathcal{F}$ over unseen training data. 

This way, the AF would be selecting its next matchup by training $\vert X_{x_t} \vert$ emulators, and taking whichever achieves the lowest error score on the remainder of our training data:
\begin{equation}\label{eq:AFcheat}\begin{split}
\text{AF}_\text{cheat}(E, T_1, T_2) = -\text{avg(err(}E', x\text{) for }x\text{ in training data)},
\end{split}
\end{equation}
where $E'$ is a copy of emulator $E$ after being trained on the outcome of a match between $T_1$ and $T_2$.

\sss{Gaussian Process}
With our EI assumptions, we can note two factors inhibiting the usefulness of a simple GP for surrogate model $s_\mathcal{F}$:

\begin{itemize}
\item \textbf{The dynamic between X and Y changes} as $t$ increments. A gaussian process could model this as variance in X, but doing so would fail to capture what should be a predictable dynamic (e.g. $\mathcal{F}_{(x_2, x_2)}(x_2)$ will almost certainly be higher than $\mathcal{F}_{(x_1, x_1)}(x_2)$).
\item \textbf{Our model cannot directly sample Y} at each iteration, as computing Y depends on unseen future matches. Instead, the model can only sample the outcome of one match per iteration, and must \textit{approximate} Y based on:
    \begin{itemize}
    \item The information content of $x_t$.
    \item An assumption about how our emulator will make use of this information content.
    \item An assumption about the distribution of matchups on which the emulator would be evaluated.
    \end{itemize}
\end{itemize}
\medskip
We can address both of the above by focusing our acquisition function's design goal into the following:

\medskip
\begin{center}
\textit{\textbf{How can we compute the information content of a particular matchup, in the context of whichever matches the emulator has seen thus far?}}
\end{center}
\medskip

\sss{Information-Theoretic AF: Likeliest Draw}

In designing an information-theoretic acquisition function, an intuitive option would be to take the matchups that our emulator believes are \textit{most likely to be draws}, in order to settle uncertainties of skill between pairs of players. We can justify this as being the entropy of an individual matchup.

Let us suppose that $R$ is a discrete random variable (DRV) representing the emulator's predicted results for a particular matchup, having two outcomes: $w_{T1}$ and $w_{T2}$. Furthermore, suppose $M$ is a DRV representing all matchups seen by the emulator. In this regard, $H(R|M=m)$ represents the entropy of the two possible results for a particular matchup according to our emulator.

We can derive a formula for the entropy in terms of the emulator's predictions $p(w_{T1}|m)$ and $p(w_{T2}|m)$:
\[\begin{split}
\theta = H(R|M=m) = -p(w_{T1}|m) \log(p(w_{T1}|m)) \\ - p(w_{T2}|m) \log(p(w_{T2}|m)).
\end{split}\]
Taking $p(w_{T2}|m)=1-p(w_{T1}|m)$:
\[\begin{split}
\theta = H(R|M=m) = -p(w_{T1}|m) \log(p(w_{T1}|m)) \\ - (1-p(w_{T1}|m)) \log(1-p(w_{T1}|m)).
\end{split}\]
We can then show that the entropy is maximized when $p(w_{T1}|m)=0.5$ by taking the derivative:
\begin{align*}
 \frac{d\theta}{dp} &= log(1 - p(w_{T1}|m)) - log(p(w_{T1}|m) \\
0 &= log(1 - p(w_{T1}|m)) - log(p(w_{T1}|m) \\
1 &= \frac{1 - p(w_{T1}|m)}{p(w_{T1}|m)}  \\
0.5 &= p(w_{T1}|m).
\end{align*}
Thus, we define acquisition function:
\begin{equation}\label{eq:AFdraw}\begin{split}
\text{AF}_\text{draw}(E, T_1, T_2) = -p(w_{T1}|m) \log(p(w_{T1}|m)) \\ - (1-p(w_{T1}|m)) \log(1-p(w_{T1}|m)).
\end{split}
\end{equation}

\sss{Information-Theoretic AF: Cross-Entropy}

The likeliest-draw approach has a key limitation: it neglects to consider the number of prior encounters that an emulator has had with a specific matchup.

The failure here is to assume that the entropy within a match is reflective of the entropy within our emulator's understanding of the world. In actual fact, some teams will naturally have a new-draw winrate against one another, and sampling these matchups many times over won't be informative to the emulator as time goes on.

To rectify this, we propose reformulating the entropy calculation to instead compute the surprisal of each result \textit{in the context of all results that the emulator has previously seen}. The assumption here is that more unexpected results will be more informative to the emulator.

This is equivalent to the Cross-Entropy (CE) between the emulator's predicted distribution of results for a particular matchup $(R|M=m)$, and its predicted distribution of results across all known matchups $(R|M)$, derived as follows:
\begin{align*}
&CE((R|M=m), (R|M)) \\
 &= -\mathbb{E}_{(R|M=m)}[\log (R|M)] \\
 &= -p(w_{T_1}|m) * \log(p(w_{T_1})) - p(w_{T_2}|m) * \log(p(w_{T_2})).
\end{align*}
By Bayes' theorem, we take $p(w_{T_1}) = p(w_{T_1}|m) p(m)$:
\begin{align*}
C&E((R|M=m), (R|M)) \\
\begin{split}
 = &-p(w_{T_1}|m) \log(p(w_{T_1}|m)p(m)) \\ &- p(w_{T_2}|m) \log(p(w_{T_2}|m)p(m))
\end{split} \\
\begin{split}
 = &-p(w_{T_1}|m) \log(p(w_{T_1}|m)p(m)) \\ &- (1-p(w_{T_1}|m)) \log((1-p(w_{T_1}|m))p(m)).
\end{split}
\end{align*}
Here, $p(m)$ represents the probability of matchup $m$ occurring according to the emulator. By assuming matchups are independently distributed between teams, we can find $p(m) = \frac{c(T1)}{\sum_{T \in Team}c(T)} \cdot \frac{c(T_2)}{\sum_{T \in Team}c(T)}$, where $c(T_1)$ is the number of times that the emulator has observed team $T_1$. This gives us a computable expression:
\begin{align} \notag
&\text{AF}_\text{CE}(E, T_1, T_2) = CE((R|M=m), (R|M))\\
\begin{split} \label{eq:AFCE}
 &= -p(w_{T1}|m) \log \left(p(w_{T_1}|m)\frac{c(T_1)}{\sum c(T)} \frac{c(T_2)}{\sum c(T)} \right) \\ &- (1-p(w_{T_1}|m)) \log \left((1-p(w_{T_1}|m))\frac{c(T_1)}{\sum c(T)} \frac{c(T_2)}{\sum c(T)}\right).
\end{split}
\end{align}

\sss{Weighted AF}

One could think of $\text{AF}_\text{CE}$ as a function of two factors: the estimated likelihood of a draw between two teams, and the number of times our emulator has seen those teams before. However, it is not clear how to parameterise these factors within $\text{AF}_\text{CE}$.

Instead, we propose a weighted acquisition function which models the two factors (draw probability, no. of times teams seen) explicitly:
\begin{align} \notag
&\text{AF}_\text{weighted}(E, T_1, T_2) = \alpha \cdot \text{draw\_factor} + \beta \cdot \text{seen\_factor} \\  \notag
&=\alpha \cdot (1-|p(w_{T_1}|m)-p(w_{T_2}|m)|) \\ \label{eq:AFweighted}
& + \beta \cdot \left( \left(\frac{1}{c(T_1)} - \frac{1}{c(T_1)+1}\right) + \left(\frac{1}{c(T_2)} - \frac{1}{c(T_2)+1}\right) \right).
\end{align}

\sss{Other Acquisition Functions} \label{sec:AFother}

For the sake of comparison and experimentation, we implemented several other simple acquisition functions:
\begin{itemize}
\item \textbf{LeastSeen}: simply sum the number of times an emulator has seen each team (or, in the case of \textit{TSPlayers}, players) and perform a negative logarithm on each result:
\begin{equation} \label{AFleastseen}
\text{AF}_\text{unseen}=-\sum_{T \in m}\log(c(T)).
\end{equation}
This is another information-theoretic approach that equates observations with bits of information, though neglects to take into account result predictions.
\item \textbf{MostSeen}: take the negative of LeastSeen. \textit{\underline{We expect this AF to perform poorly}, by the logic that more expected matchups will tend to have lesser information content for the Emulator.}
\item \textbf{LikeliestWin}: take the negative of LikeliestDraw (\ref{eq:AFdraw}). \textit{\underline{We expect this AF to perform poorly}, by the logic that more ``obvious" outcomes will also convey lesser information.}
\item \textbf{TSQuality}: take TrueSkill's built-in \texttt{quality()} metric. The TrueSkill paper describes this as ``\textit{the draw
probability relative to the highest possible draw probability in the limit $\epsilon\rightarrow0$}", which in practice can be thought of as the expectation of draw probability when treating team skill as a Gaussian Process. In other words, a trade-off between an Emulator's confidence in player skill and its confidence in match outcome.
\end{itemize}

\subsection{Data} \label{sec:data}

\maintable{t}

For our experiments, we scraped a large database of professional and semi-professional CS:GO matches from \href{https://hltv.org}{hltv.org}. For the analysis, we used all matches with a greater than or equal to ``1 star'' team rating between 2017-2022, for a total of 9,929 matches.  %

As the modelling of skill ratings over time is out of scope of this work, we split the dataset into a training and evaluation set using a random 50/50 split. Matches used for fitting by the AFs are taken from the training set, and evaluation is always done by analysing emulator performance on the entire validation set.

\subsection{Trueskill Sensitivity Analysis}

Algorithms such as TrueSkill naturally have several parameters that control their behaviour and thus performance (in the case of Trueskill, there are four primary parameters as described above: $\mu, \sigma, \beta, \tau$). While the defaults are generally considered to ``lead to reasonable dynamics" \cite{trueskillmath}, there is no current literature exploring the selection of these parameters.

Hence, we here perform a sensitivity analysis against our dataset to determine which parameters have the largest effect on performance, and whether the defaults provide reasonable results on our dataset. Since all parameters are set relative to $\mu$, we perform a logarithmic grid search across each pair of parameters in $\sigma, \beta, \tau$, $\pm 1$ order of magnitude from the default.

Since all other parameters are defined in terms of $\mu$, we leave the default $\mu=25$, and vary two parameters at a time out of $\sigma, \beta, \tau$ in the form of a logarithmic grid search. We then use the \textbf{GPy} Python library \cite{gpy}, to fit a Gaussian Process with an RBF kernel to the results:
$$k(x, x') = \frac{1}{1000} \exp\left(-\frac{||x - x'||^2}{0.5}\right)$$
with a mean performance prior of $60\%$. Using a GP allows us to smooth the noisy results without needing to repeat every run many times to get an average.

\pagebreak

\section{Results \& Analysis}

\sensitivityanalysisfigure{t}

\accuracytraininggraph{t}

Table \ref{table:main} summarises the results across each combination of emulator and acquisition function. As expected, we see that as more matches are selected to fit the emulator, the overall performance of that emulator increases. 

The Weighted acquisition function (parameterised with $\alpha=1, \beta=1$) introduced in this work produces the best overall performance for all emulators based on existing rating systems, with the biggest gains seen after only 500 matches of training. The Random AF provides a baseline to which we can compare the other acquisition functions; we see that the MostSeen and LikeliestWin AFs provide significantly worse performance for the same number of training matches. This is inline with theory, as we are essentially feeding the emulator the results of matches it is already sure about. However, this highlights the importance of choosing a good acquisition function when minimal data is available.

The LikeliestDraw and CrossEntropy AFs select matchups which the emulator is unsure about, and this yields an improvement of 1-1.5\% over random matchups from the dataset. However, we see that while the Weighted AF consistently provides the best performance for Elo, The more basic WinRate emulator favours the CrossEntropy AF. 

The TSQuality AF, which uses TrueSkill's own internal match quality for selection, underperforms in comparison to other AFs, even when compared to random sampling.

We can also analyse the comparative behaviour of different Emulators. We find that among the team-based skill rating systems, Glicko2 provides better performance across the board, with greater gains with fewer training matches. This is surprising, as TrueSkill was introduced to address the shortcomings of Glicko2.

However, one of the main benefits that TrueSkill brings is the ability to generalise beyond 1v1 matchups, and the benefit of this is seen when we use 5v5 matchups in TrueSkill with per-player skill ratings. The TSPlayers emulator beats all other approaches by around 1\%, giving us the best achieved average accuracy of 64.1\%.

Overall, we see that the choice of acquisition function and emulator can be largely decoupled: player-based TrueSkill excels as an emulator throughout, and the Weighted acquisition function provides the best results across emulators.

\pagebreak

\subsection{Trueskill Sensitivity Analysis}

Figure \ref{fig:ts-sensitivity} shows us the performance of each of our TrueSkill-emulators, as the algorithm parameters are varied. We can draw a number of interesting conclusions from this:

\begin{enumerate}
\item $\beta$ and $\sigma$ seem like the most important parameters, with $\tau$ having a much smaller effect on performance. We confirm that the ratio between $\beta$ and $\sigma$ is very important ($\beta = \sigma / 2$ is the default) - performance remains similar if the ratio is maintained. We find that the optimal ratio is $\beta = \sigma / 0.5$ and $\beta = \sigma / 1.6$ for the two emulators.
\item In \textbf{TrueSkillEmulator} (with a single skill rating for a given team), the combination of a high $\sigma$ and low $\beta$ dramatically reduces the performance of the algorithm. This effect is also seen (less drastically) in \textbf{TrueSkillPlayersEmulator}, which maintains a separate skill rating for each of the team's 5 players.
\item The \textbf{TrueSkillPlayersEmulator} is much more robust to changes in parameters (a 1.3\% range versus 7.5\%). This suggests that the TrueSkill algorithm is more stable with 5v5 matches rather than 1v1 matches.
\item That being said, we see that the default values given achieve close to optimal performance in both cases! We can make slight gains in the per-team emulator, but the cost of modifying parameters in the wrong direction is much greater.
\end{enumerate}

\subsection{Accuracy over training}

Figure \ref{fig:curve} shows the performance of emulators using the Random and Weighted acquisition function, and the emulators are trained until they run out of matches.

We find that the engineered AFs reach a plateau of performance much faster than picking matches at random: matches are picked that try to maximise the speed at which the emulators can learn. In the case of the Weighted AF, we actually see overall accuracy actually decreases towards the end of training. This is because a good acquisition function selects ``high-quality'' matches (for some notion of high quality), which inevitably means that the `low-quality' matches are all fed to the emulator at the end; potentially an unrepresentative sample that skews ratings. We see that this behaviour is much more prominent in the TrueSkillPlayers emulator.

Note that by intentionally training on an AF-selected subset of the dataset, we avoid this behaviour in  Table \ref{table:main}, as we report results there only after training on a subset of the dataset. This effect is a limitation of our training method, which limits matchups to a subset of actually observed matchups (rather than letting the model choose any possible matchup), as opposed to a fundamental property of the emulators.

Overall, the amount of variance observed run-to-run was a surprising result. This means that on any given run (especially with the Random AF), any two skill rating systems could achieve comparable performance - however, there are significant differences between \textit{average} performance across many runs.

\section{Discussion}

\subsection{Limitations} \label{sec:limitations}

While the goal of this work was to accurately model each method's performance within an environment where it can choose which match is played next, there are several limitations to our work:
\begin{itemize}
\item Not every matchup is selectable by the emulator during training - it can only choose matchups that actually occurred in the dataset.
\item Matches are not presented to the algorithm in \textit{time-order}, meaning that the ability of each emulator to model \textit{time-varying} skill was not tested. This was done to avoid dramatically limiting match choice.
\item We test only on a single dataset, based on professional CS:GO matches. Therefore, it is unclear how well the results transfer to amateur matches, or other video games which rely on similar matchmaking systems.
\item We have only evaluated Emulators in their accuracy at predicting non-draw outcomes. It may be insightful to compute the log-loss of each emulator to reward stronger beliefs \& to factor in drawed games.
\item Acquisition Functions are evaluated on their ability to produce accurate skill ratings (as measured by match prediction accuracy), but there are many other metrics. For example, the TrueSkill Quality metric \cite{trueskill} (which did not perform well in this work) is designed to try to maximise the probability of a draw, under the heuristic that evenly matched games are the most fun; the best match for players may not be the match that allows the skill rating system to learn the most.
\end{itemize}

\subsection{Future Work} \label{sec:futurework}

Perhaps the primary way in which our work could be extended is through the use of larger datasets (such as amateur matchmaking data if such data were obtainable), which would allow for greatly reduced aleatoric uncertainty in results. Another obvious extension would be to analyse several different games and determine whether the choice of system is game-specific or if e.g. TrueSkill is always superior to Elo.

Other avenues include:
\begin{itemize}
\item \textbf{Parameter-tuned acquisition functions.} To further improve the performance of our Weighted AF (\ref{eq:AFweighted}), we could consider tuning its parameters \textit{per emulator}, such as through Bayesian optimisation. Furthermore, we could explore the use of highly parameterized AFs that make deeper assumptions about the emulator and data, such as modeling the impact of network effects on different teams; or even the relationship between factors like draw probability and team count as a Gaussian process to achieve closer to optimal results.
\item \textbf{Exploring non-tournament matchmaking settings.} To provide a more realistic matchmaking experience, we could simulate a setting whereby the matchmaking function must balance the usefulness of the emulator, fairness of the match, and queue waiting times, rather than assuming that all teams are always available for matches. This approach would offer a more dynamic and practical solution.
\item \textbf{Next-generation emulators}. Microsoft recently published TrueSkill2 \cite{trueskill2}, which claims to improve match prediction accuracy on Halo 5 from 52\% to 68\%. This was out of scope in this paper as no open-source implementations currently exist. Alternatives that could be evaluated include OpenSkill. \cite{openskill}.
\end{itemize}

\section{Conclusion}

In this work, we performed an thorough analysis on the performance of different skill rating systems and matchmaking algorithms when applied to a real-world dataset of professional CS:GO matches. The core novelty of our work is the use of a surrogate modelling architecture to evaluate a skill rating system's effect on matchmaking and its recursive effect on future skill ratings, while retaining the use of real-world data. Our architecture is released as a Python library to allow for future extension and evaluation.

We draw several conclusions from our analysis: that the default parameters of TrueSkill yield close to optimal results, that a selection of emulator and AF can be made largely independently, that a 5v5 TrueSkill and Weighted AF provide the best results, and that optimal performance can be reached with surprisingly few matches, with only marginal gains between skill rating systems. Larger datasets and an evaluation of generalisation across games could allow for more robust future analysis.

\section{Source Code}

Our \textbf{skillbench} framework, including datasets and all the emulators used in this work is available at \url{https://github.com/mgm52/skillbench}.

\bibliographystyle{IEEEtran}
\bibliography{bibliography}

\clearpage

\end{document}